\documentclass[conference]{IEEEtran}
\IEEEoverridecommandlockouts
\usepackage{amsmath,amssymb,amsfonts}
\usepackage{graphicx,bm}
\usepackage{multirow}
\usepackage{algorithm}
\usepackage{algorithmic}
\newcommand{\sg}{\mbox{sgn}}

\begin{document}

\title{
Analysis and FPGA based Implementation of Permutation Binary Neural Networks
}

\author{\IEEEauthorblockN{Mikito Onuki, Kento Saka and Toshimichi Saito}
\IEEEauthorblockA{EE Dept., HOSEI University, Koganei, Tokyo, 184-8584 Japan\\
mikito.onuki.8v@stu.hosei.ac.jp, \ kenken3731127@ezweb.ne.jp, \  tsaito@hosei.ac.jp}
}
\maketitle

\begin{abstract}
This paper studies a permutation binary neural network characterized by local binary connections, global permutation connections, and the signum activation function.
Depending on the permutation connections, the network can generate various periodic orbits of binary vectors. 
Especially, we focus on globally stable periodic orbits such that almost all initial points fall into the orbits. 
In order to explore the periodic orbits, 
we present a simple evolutionary algorithm. 
Applying the algorithm to typical examples of PBNNs, 
existence of a variety of periodic orbits is clarified. 
Presenting an FPGA based hardware prototype, 
typical periodic orbits are confirmed experimentally. 
The hardware will be developed into various engineering applications
such that stable control signals of switching circuits 
and stable approximation signals of time-series. 
\end{abstract}

\section{Introduction}
\label{intro}
Discrete-time recurrent neural networks (DT-RNNs \cite{rnn1} \cite{rnn2}) are characterized by nonlinear activation function and real valued connection parameters. 
The dynamics is described by an autonomous difference equation of real state variables. 
Depending on the parameters, the DT-RNNs can exhibit various interesting phenomena: periodic orbits, chaos, and related bifurcation. 
The real/potential applications include
dynamic associative memories \cite{rnn2}, 
combinatorial optimization problems solvers \cite{tsp}, and 
reservoir computing \cite{rc}. 
The DT-RNNs are important subject in analysis of nonlinear dynamics and engineering applications. 
An important basic problem is relationship between parameters and existence/stability of periodic orbits. 
However, this problem is not easy because of complicated dynamics for an enormous number of parameters. 
In order to consider the problem precisely, a simple DT-RNN with various periodic orbits is essential. 

This paper studies a simple DT-RNN: a permutation binary neural network (PBNN) characterized by local binary connections, global permutation connections, and the signum activation function \cite{taka} \cite{dcdsb}.
The dynamics is described by an autonomous difference equation of binary state variables.
The PBNN is well suited for precise analysis of periodic orbits and FPGA based hardware implementation. 
Although the number of parameters is much smaller than DT-RNNs with real values connection parameters, 
the PBNNs can generate various periodic orbits of binary vectors (BPOs). 
Especially, we focus on globally stable binary periodic orbits (GBPOs) 
such that almost all initial points fall into the GBPOs. 
Real/potential applications of the GBPOs include
control signals of switching circuits \cite{pe1} \cite{pe2} 
and approximation signals of time-series \cite{rc}. 
Because of the global stability, the signals are robust and have error correcting function \cite{error}.

First, in order to analyze the PBNN, we give basic definitions: 
the difference equation of the PBNN, 
global stability of BPOs, and 
canonical permutation identifier (CPID) for classification of the global permutation connections,  
The CPID corresponds to key parameter in the analysis. 

Second, referring to evolutionary algorithms \cite{ec1} \cite{ec2}, 
we present a simple algorithm to explore GBPOs. 
In the algorithm, individuals correspond to CPIDs 
and an operator corresponds to exchange of permutation connection wires, 
The individuals are evaluated by stability of BPOs and 
the offspring are generated by elitism. 
Applying the algorithm to typical examples of PBNNs, 
existence of a variety of GBPOs is clarified. 
Note that the brute force attack is possible to find all the GBPOs in low dimensional PBNN, however, the computation cost increases as the dimension increases.  
Our algorithm can find typical samples of GBPOs in lower computation cost.  

Third, we present a simple FPGA based hardware prototype and 
typical GBPOs are confirmed experimentally. 
The hardware is based on simple Boolean functions and 
is efficient in power consumption. 
The hardware will be developed into engineering applications.

\section{Permutation binary neural networks}
\begin{figure}[t!]
\centering
\includegraphics[width=1\columnwidth]{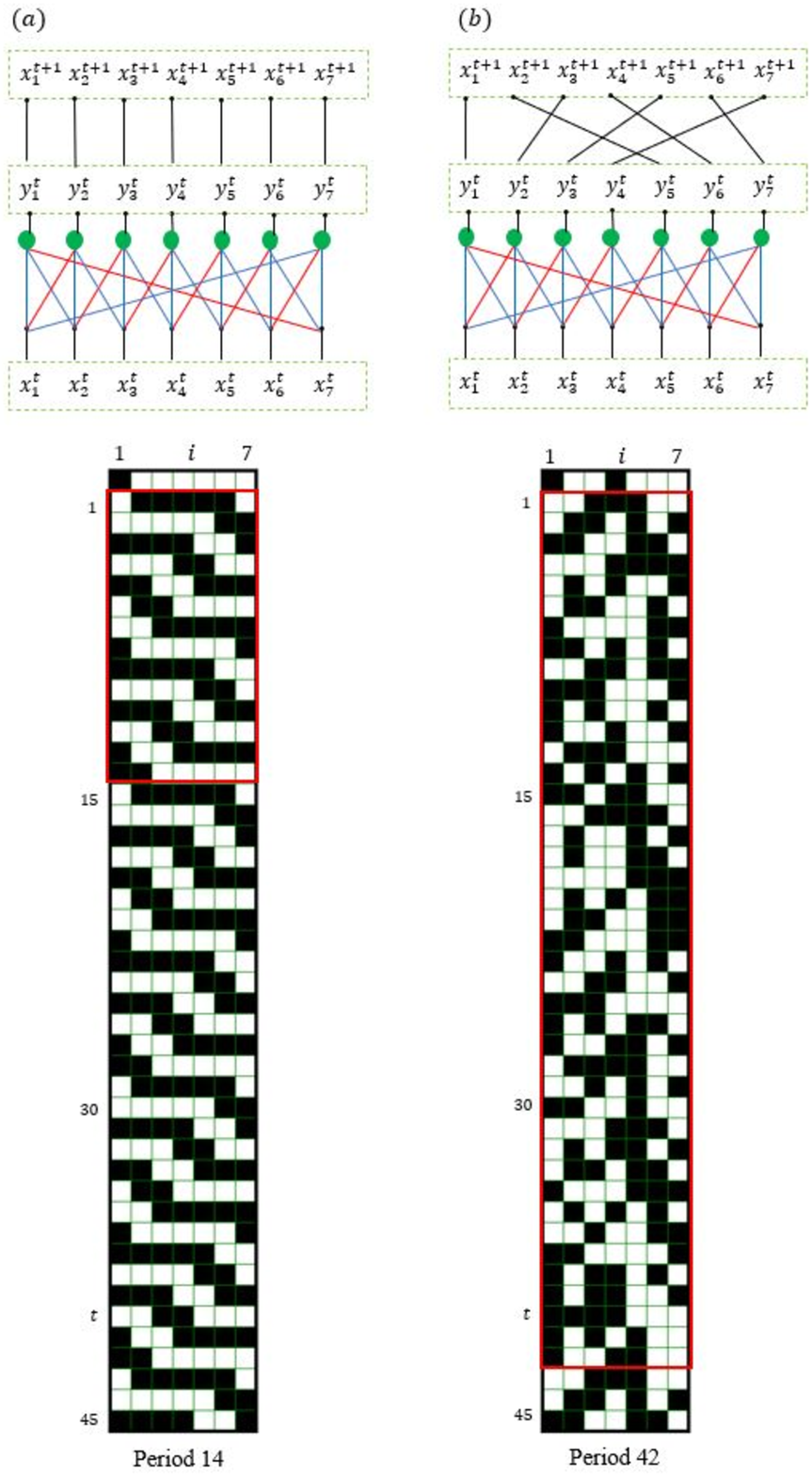}
\caption{
PBNN and BPO.
(a) $P(1234567)$, BPO with period 14. 
(b) $P(1526374)$, BPO with period 42. 
}
\label{fg1} 

\vspace*{5mm}

\centering
\includegraphics[width=1\columnwidth]{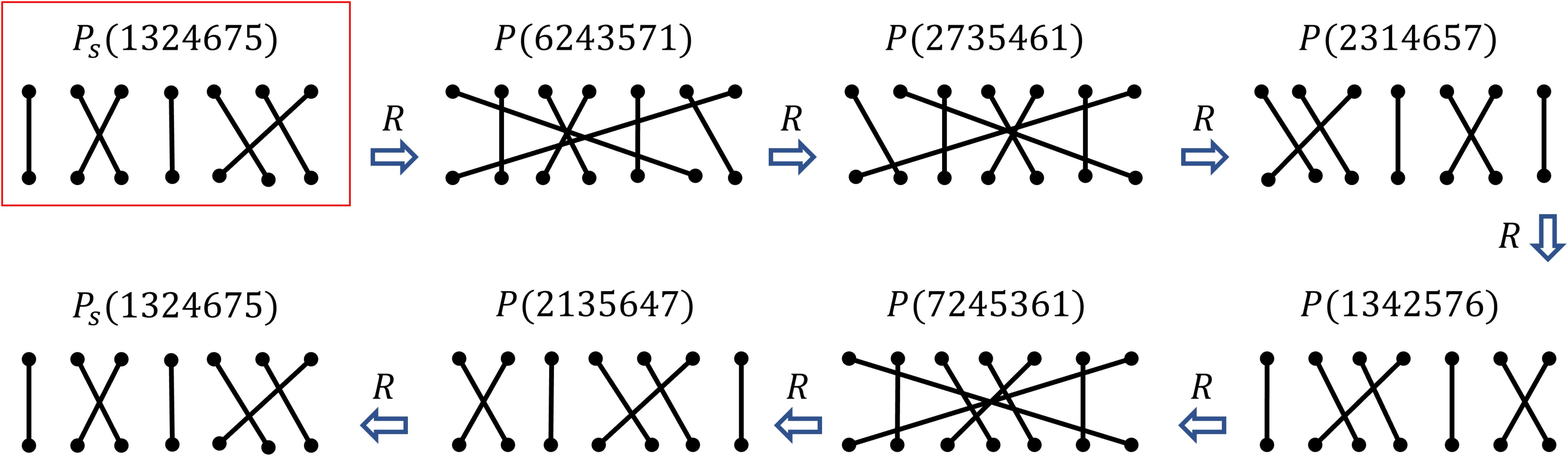}
\caption{
Canonical permutation identifier (CPID) $P_s$ and equivalent global permutation connections. 
}
\label{fg2} 
\end{figure}

PBNN dynamics is described by $N$-dimensional autonomous difference equation of 
binary state variables:
\begin{equation}
\begin{array}{c}
x_i^{t+1} = y_{\sigma(i)}^t, \ 
y_i^t = \sg \left(w_a x_{i-1}^t + w_b x_i^t + w_c x_{i+1}^t \right)\\
\sigma = \left(
    \begin{array}{cccc}
      1         & 2         & \cdots & N \\
      \sigma(1) & \sigma(2) & \cdots & \sigma(N)
    \end{array}
  \right)\\
\sg(x) = \left\{
\begin{array}{lll}
+1 & \mbox{if } x \ge 0, & \ i \in \{1, \cdots, N \} \\
-1 & \mbox{if } x < 0,   & \ j \in \{1, \cdots, M \}
\end{array}\right. 
\end{array}
\label{pbnn}
\end{equation}
where
$x_i^t \in \bm{B}$ is the $i$-th binary state variable at discrete time $t$,  
$y_i^t \in \bm{B}$ is the $i$-th binary hidden state, and $\sigma$ is a permutation. 
$x_0^t \equiv x_N^t$ and $x_{N+1}^t \equiv x_1^t$ for ring-type connection as shown in Fig. \ref{fg1}.  
Let $\bm{x}^t \equiv (x_1^t, \cdots, x_N^t)$ and let $\bm{y}^t \equiv (y_1^t, \cdots, y_N^t)$.
The local binary connection (from input to hidden layers) 
transforms a binary input vector $\bm{x}^t$ into the binary hidden vector $\bm{y}^t$.
The global permutation connection (from hidden to output layers) transforms the $\bm{y}^t$ into the binary output vector $\bm{x}^{t+1}$. 
The output $x_i^{t+1}$ is fed back to the input layer and 
the PBNN can generate various sequences of binary vectors. 
For convenience, Eq. (\ref{pbnn}) is abbreviated by 
\[
\bm{x}^{t+1} = F(\bm{x}^t) 
\]
The local binary connections are identified by the connection number
\[
\begin{array}{cc}
\mbox{CN0}: \bm{w}_l=(-1, -1, -1) & 
\mbox{CN1}: \bm{w}_l=(-1, -1, +1) \\ 
\mbox{CN2}: \bm{w}_l=(-1, +1, -1) &
\mbox{CN3}: \bm{w}_l=(-1, +1, +1) \\ 
\mbox{CN4}: \bm{w}_l=(+1, -1, -1) &
\mbox{CN5}: \bm{w}_l=(+1, -1, +1) \\ 
\mbox{CN6}: \bm{w}_l=(+1, +1, -1) &
\mbox{CN7}: \bm{w}_l=(+1, +1, +1) 
\end{array}
\]
The global permutation connections are identified by 
\[
\mbox{Permutation identifier: }  P(\sigma(1) \cdots \sigma(N)). 
\]
Fig. \ref{fg1} shows examples of 7-dimensional PBNNs for CN1. 
For identity permutation P(123456), the PBNN exhibits a BPO with period 14. 
Applying permutation P(1526374), the PBNN exhibits a BPO with longer period 42. 
Here we give several basic definitions. 

{\it Definition 1}: 
A point $\bm{z}_p \in \bm{B}^N$ is said to be a binary periodic point (BPP) with period $p$ if 
$\bm{z}_p = F^p(\bm{z}_p)$ and 
$F(\bm{z}_p)$ to $F^p(\bm{z}_p)$ are all different where 
$F^k$ is the $k$-fold composition of $F$. 
A sequence of the binary periodic points 
$\{F(\bm{z}_p ), \cdots, F^p(\bm{z}_p) \}$ 
is said to be a binary periodic orbit (BPO) with period $p$. 
A point $\bm{z}_e \in \bm{B}^N$ is said to be an eventually periodic point (EPP) of a BPO 
if $\bm{z}_e$ is not a BPP but falls into the BPO. 

{\it Definition 2}: 
A BPO is said to be a globally stable binary periodic orbit (GBPO) if 
all initial points fall into the GBPO, except for the two endpoints:  
$\bm{x}_- \equiv (-1, \cdots, -1)$ and $\bm{x}_+\equiv(1, \cdots, 1)$. 
The endpoint is either a fixed point or a periodic point with period 2. 

{\it Definition 3}:
Let $R$ be a shift operator such that 
\begin{equation}
\begin{array}{l}
R: P_0(\sigma_0(1) \cdots \sigma_0(N_p)) \rightarrow P_1(\sigma_1(1) \cdots \sigma_1(N))\\
P_1 = R(P_0), \sigma_1(i+1) = \sigma_0(i) + 1 \mbox{ mod }  N
\end{array}
\end{equation}
where $i \in \{1, \cdots, N_p\}$ and $\sigma_1(N + 1) \equiv \sigma_1(1)$. 
Since the neurons are ring-type connection, 
the permutation connections $P_1$ and $R(P_1)$ are equivalent even if the permutation identifiers are different. 
Let $S$ be a set of permutation identifiers that give equivalent permutation connections. 
The set $S$ is represented by a canonical permutation identifier (CPID) $P_s$  
that is the minimum element in $S$ by means of base-$N$ number:
\[
P_s(\sigma_s(1) \cdots  \sigma_s(N)) < P_k(\sigma_k(1) \cdots  \sigma_k(N)) \in S,  
k \ne s
\]
Fig. \ref{fg2} shows equivalent permutation connections and their CPID.

Real/potential applications of the GBPOs include 
control signals of switching circuits and 
approximation signals of time series.  
Because of the global stability, 
the signals are robust and have an error correction function. 
Note that linear feedback shift register \cite{lfsr} can generate M-sequences (BPOs with the maximum period), however, 
the BPO includes all the points in binary state space and cannot define global stability: the M-sequences is different category from the GBPOs.

For simplicity, we consider the case where $N$ is a prime number $N_p$ hereafter. 
The main objective problem is exploring GBPOs: 
exploring CPIDs and CNs for PBNNs having GBPOs. 

\section{Evolutionary Algorithm for exploring GBPOs}

In this section, we present an evolutionary algorithm for exploring GBPOs. 
The algorithm evolves $M$ individuals corresponding to CPIDs: 
\begin{equation}
P^g = \{P_1^g, \cdots, P_M^g \}, \ P_i^g=(\sigma^g_i(1), \cdots ,\sigma^g_i(N))
\end{equation}
where $P_i^g$ denotes the $i$-th individual at generation $g$. 
The set $P^g$ of the individuals are referred to as a population. 
As a CN is given, each individual gives the CPID that determines a PBNN. 
The PBNN generates a BPO and 
the BPO is evaluated by an objective function: 
\begin{equation}
\begin{array}{l}
F_1(P^g_i) = (\#\mbox{BPPs} + \#\mbox{EPPs})/2^{N}\\
1/2^N \le F_1(\sigma_s) \le (2^N-2)/2^N \equiv F_{max}.
\end{array}
\end{equation}
If the PBNN generates multiple BPOs, we adopt one BPO that maximizes $F_1$. 
PBNN of $P_i^g$ generates a GBPO if $F(P_i^g)=F_{max}$.
An external population (EP) is used to store individuals that generate GBPOs.
The algorithm consists of two Parts: \\

\noindent
{\bf Part 1} for discovery of GBPO(s)

\noindent
{\bf Step 1:} Initialization. 
Let $g=0$ and let $\mbox{EP}=\emptyset$.
We select one connection number $CNx$.  
As an initial population $\{P^g_1, \cdots, P^g_M \}$,  
$M$ individuals are selected randomly.

\noindent
{\bf Step 2:} Evaluation. 
Each individual is evaluated by $F_1$.
If $F_1=F_{max}$ for some individual $\hat{P}^g_k$, 
then the individual gives a PBNN that generates a GBPO (discovery of a GBPO). 
The individual $\hat{P}^g_k$ is stored into the EP and go to Part 2. 

\noindent
{\bf Step 3:} Mutation. 
Applying a mutation operator to each individual, 
offspring candidates $\{\hat{P}^g_1, \cdots, \hat{P}^g_M \}$ are generated.
The mutation operator exchanges two elements selected randomly in $P$:
%
\begin{equation}
\sigma_i^g(j) \longleftrightarrow  \sigma_i^g(k) \mbox{ for  $j \ne j$ selected randomly}
\label{mut}
\end{equation}

\noindent
{\bf Step 4:} 
Let $g \gets g+1$, go to Step 2, 
and repeat until the maximum generation $g=g_{m1}$. 
If no GBPO is discovered at $g=g_{m1}$ then the algorithm is terminated. \\

\noindent
{\bf Part 2} for storage of GBPOs

\noindent 
{\bf Step 1:} Initialization. 
Let $g=0$ and let $\mbox{EP}$ consist of individuals from Step 2 in Part $1$.  
An initial population $P = \{P^g_1, \cdots, P^g_M \}$ is the same as 
the population from Step 2 in Part 1 
where element(s) can generate GBPOs. 

\noindent
{\bf Step 2:} Mutation. 
The objective population is $P_0 = P \cup P_e = \{P'^g_1, \cdots, P'^g_{M_1} \}$. 
If \#EP$ \le M_e$ then $P_e = $EP. 
If \#EP$> M_e$ then $P_e$ consists of $M_e$ elements selected randomly from the EP. 
Applying the mutation operator in Eq (\ref{mut}) to each individual in $P_0$,
offspring candidates $\{\hat{P}^g_1, \cdots, \hat{P}^g_{M_1} \}$ are generated and 
 each individual is evaluated by $F_1$.
If $F_1=F_{max}$ for some individual $\hat{P}^g_k$ then 
the corresponding PBNN generates a GBPO and 
the individual $\hat{P}^g_k$ is stored into EP. 

\noindent
{\bf Step 3:} Elitism. 
Applying the elitism to offspring candidates,
top $M$ individuals construct the next population $P=\{\hat{P}^g_1, \cdots, \hat{P}^g_M \}$.

\noindent
{\bf Step 4:}
Let $g \gets g+1$, go to Step 2, and repeat until $g=g_{max}$.\\

\noindent
We have applied the algorithm to the following examples.
\[
  \begin{array}{l}
    \mbox{Connection number: } CN1\\
    \mbox{Dimension of PBNN: } N_p=17\\
    \mbox{Individual size in Part 1: } M=50\\
    \mbox{Upper limit of generation in Part 1: } g_{m1}=1000\\
    \mbox{Additional individal size in Part 2: } M_e=50\\
    \mbox{The maximum generation in Part 2: } g_{max}=1000
  \end{array}
\]

\begin{figure}[tb]
\centering
\includegraphics[width=0.65\columnwidth]{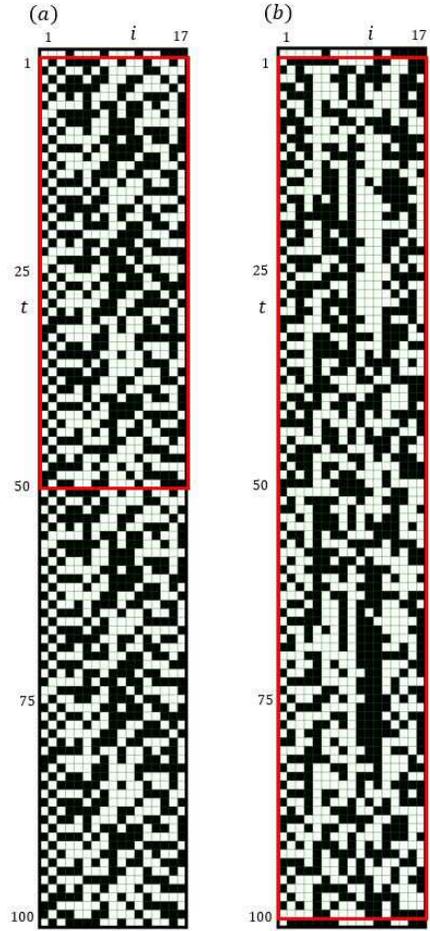}
\caption{
GBPO examples. 
(a) $P(1\: 2\: 4\: 10\: 11\: 3\: 7\: 12\: 8\: 14\: 16\: 5\: 15\: 9\: 17\: 6\: 13)$, GBPO with period 50. 
(b) $P(1\: 3\: 11\: 14\: 4\: 13\: 8\: 15\: 12\: 7\: 16\: 10\: 5\: 17\: 6\: 2\: 9)$, GBPO with period 100. 
}
\label{fg4} 
\end{figure}

\begin{figure}[tb]
\centering
\includegraphics[width=0.7\columnwidth]{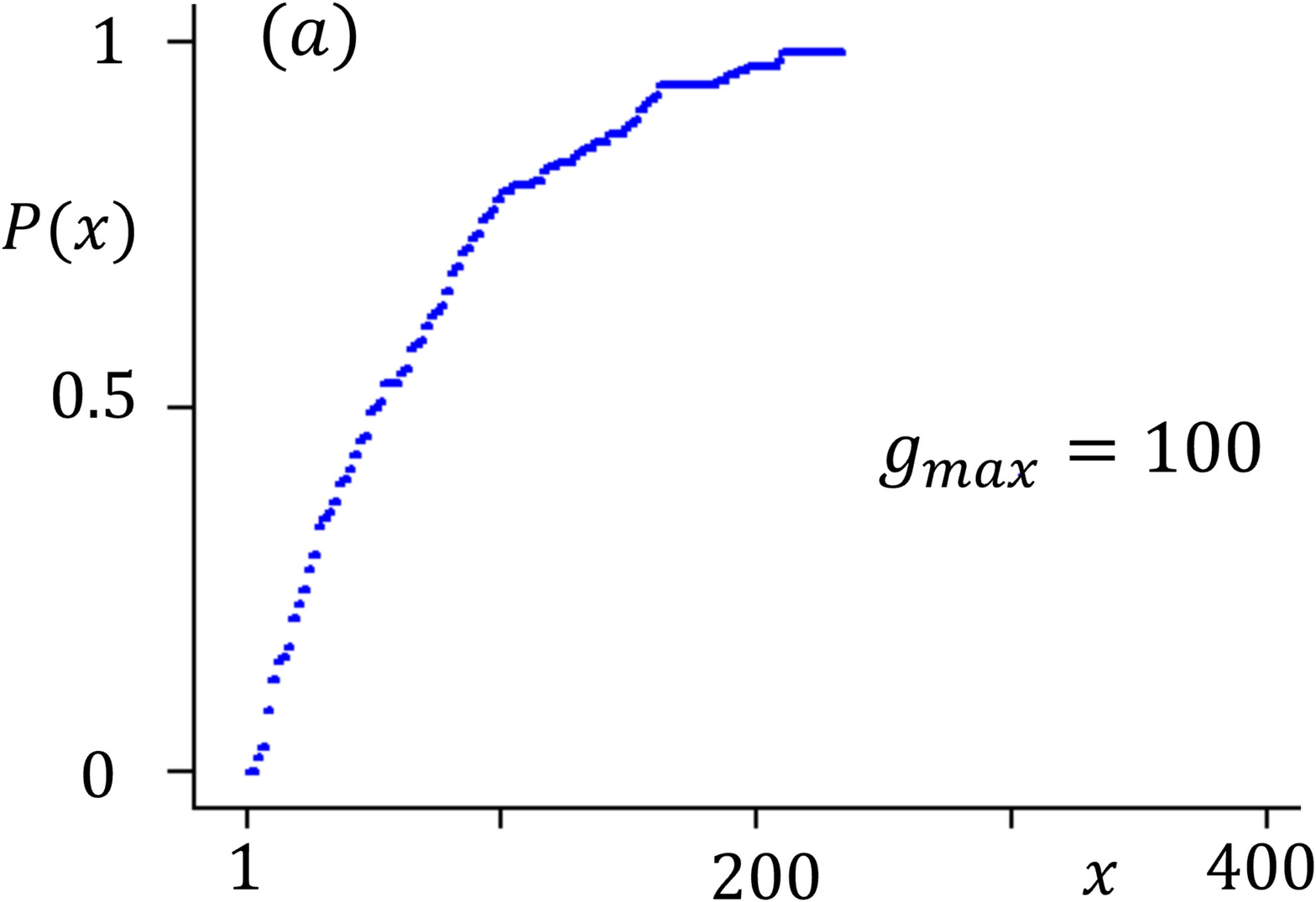}
\includegraphics[width=0.7\columnwidth]{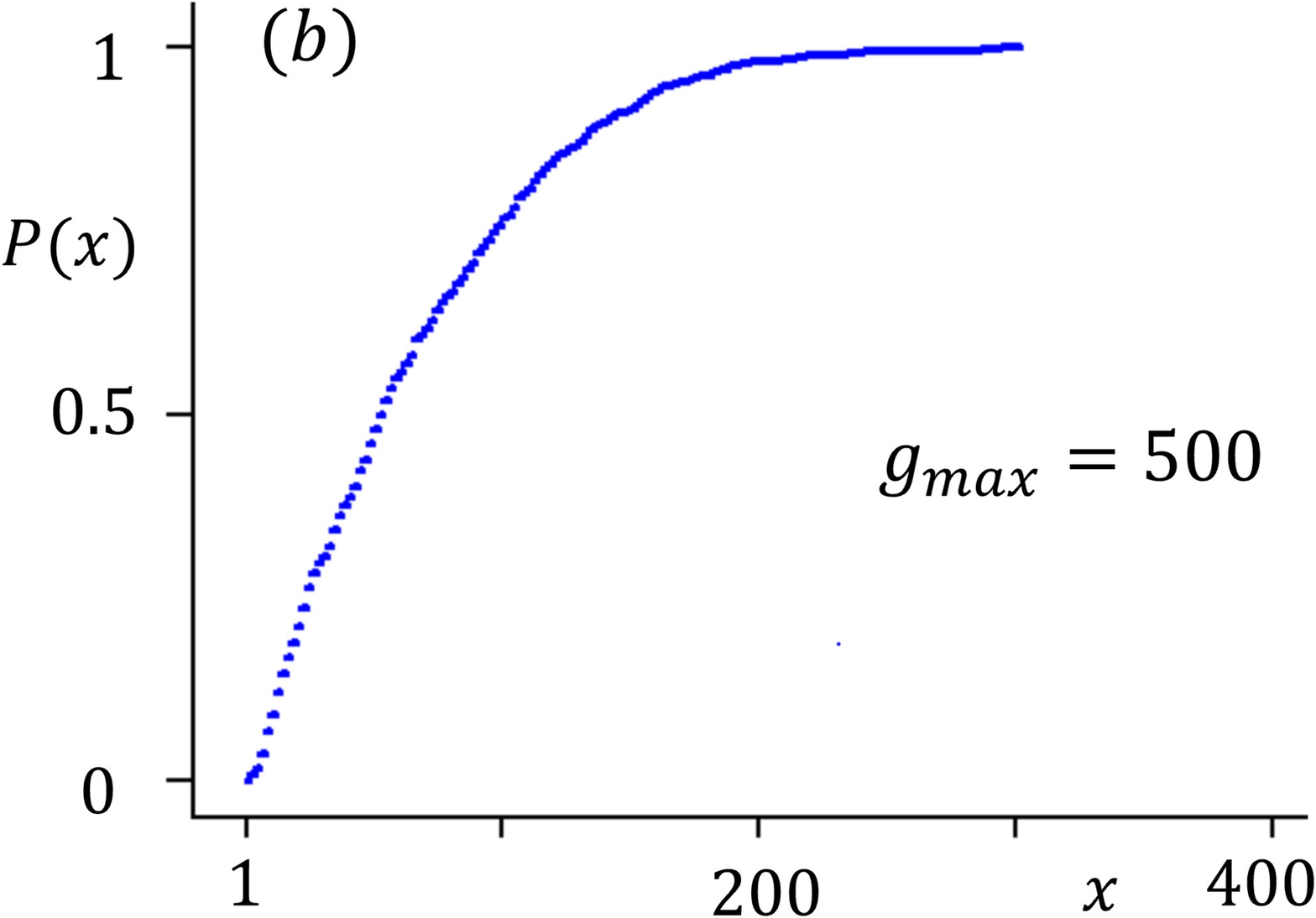}
\includegraphics[width=0.7\columnwidth]{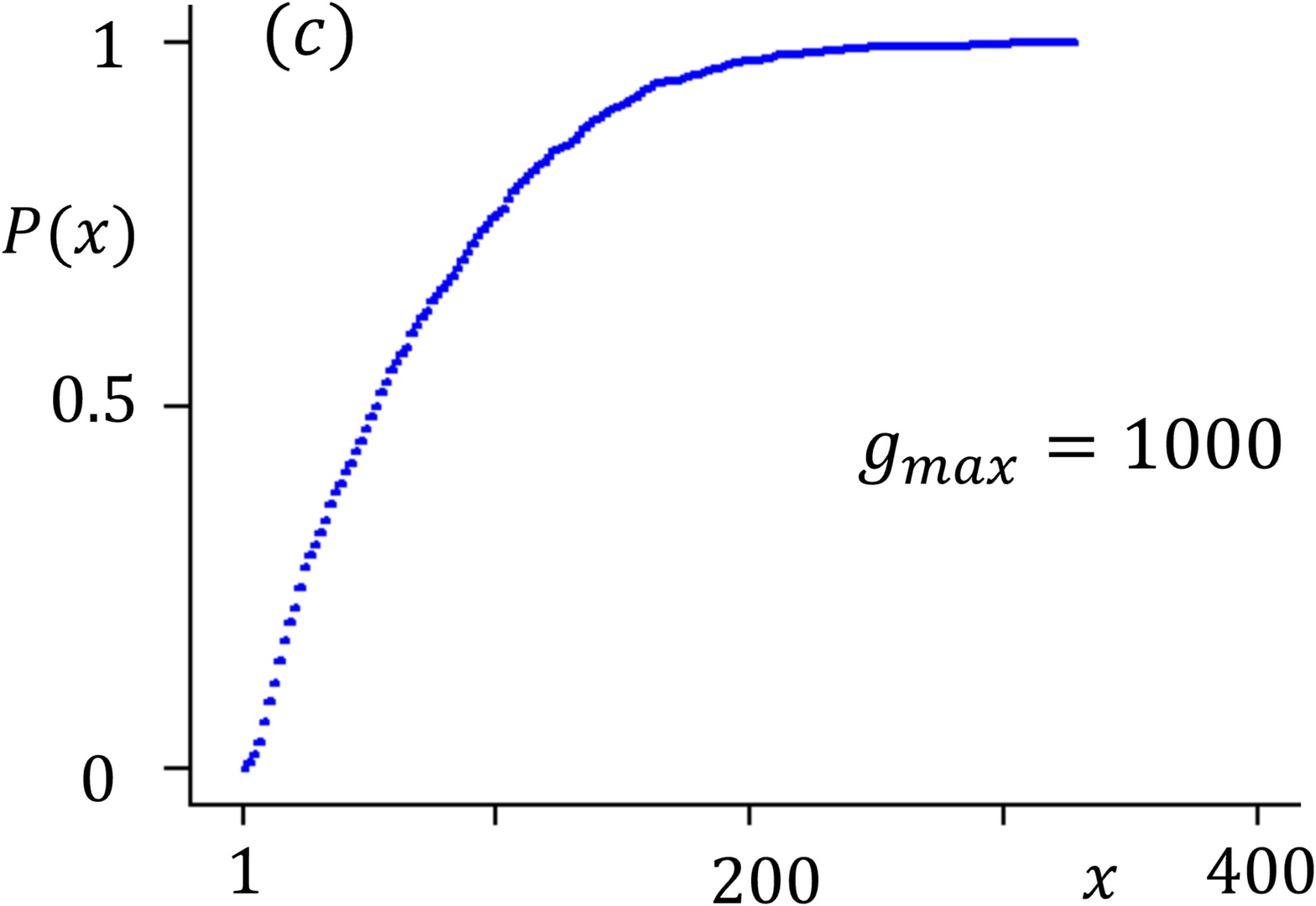}
\caption{
Cumulative distribution of period $(x)$ of GBPOs. 
$(a) g_{max}=100, P_{max}=234$
$(b) g_{max}=500, P_{max}=302$
$(c) g_{max}=1000, P_{max}=328$
}
\label{fg5} 
\end{figure}

The algorithm has discovered GBPO(s) in Part 1 and 
has found various GBPOs in Part 2. 
Two typical examples are shown in Fig. \ref{fg4}. 
As $g_{max}$ increases, the number and periods of GBPOs increase. 
Fig. \ref{fg5} shows cumulative distribution of periods ($x$) of GBPOs for several values of $g_{max}$.  
As $g_{max}$ is not large enough, shape of the distribution fluctuates. 
As $g_{max}$ is large (e.g., $g_{max}=500$), the shape converges sufficiently. 
The cumulative distribution variety shows that the PBNN can generate a variety of GBPOs. 

\section{FPGA based hardware implementation}

\begin{figure}[tb]
\centering
\includegraphics[width=0.9\columnwidth]{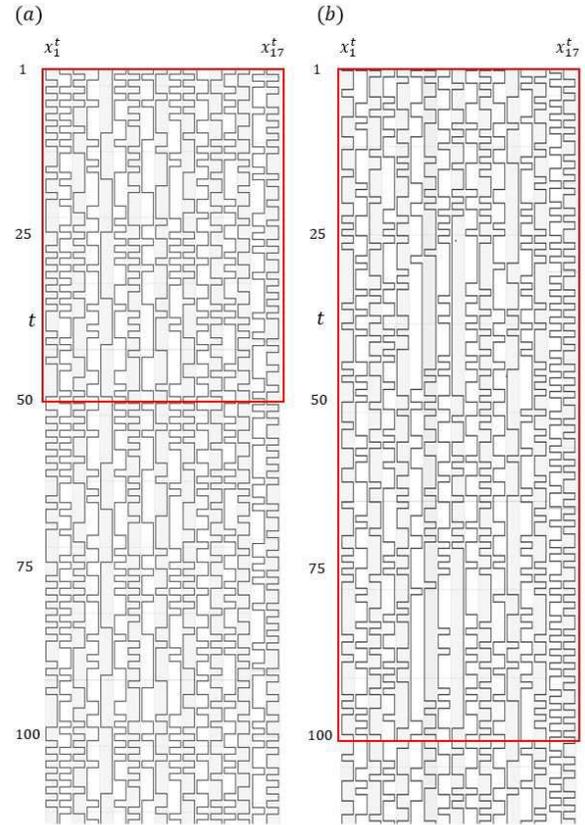}
\caption{
Measured waveforms of GBPOs on the FPGA based hardware prototype.
(a) $P(1\: 2\: 4\: 10\: 11\: 3\: 7\: 12\: 8\: 14\: 16\: 5\: 15\: 9\: 17\: 6\: 13)$, GBPO with period 50. 
(b) $P(1\: 3\: 11\: 14\: 4\: 13\: 8\: 15\: 12\: 7\: 16\: 10\: 5\: 17\: 6\: 2\: 9)$, GBPO with period 100. 
}
\label{fg6} 
\end{figure}

In this section, we present an FPGA based hardware prototype of the PBNNs. 
The FPGA is an integrated circuit designed to be configured by a designer after manufacturing.   
The advantages include high speed/precision operation and high degree of integration. 
We have designed/fabricated the hardware prototype in the following environment. 
\begin{itemize}
\item Vivado version: Vivado 2020.1 platform (Xilinx).
\item FPGA:  Xilinx Artix-7 XC7A35T-ICPG236C.
\item Clock: 1 [kHz]. The default frequency 100[MHz] is divided for clear measurement. 
\item Measuring instrument: ANALOG DISCOVERY2. 
\item Multi-instrument software: Waveforms 2015
\end{itemize}
Algorithms \ref{alg1} and \ref{alg2} show SystemVerilog codes for the design. 
In the Algorithm \ref{alg1}, as a connection number CN is given, the local binary connection is realized.   
This design strategy is based on simple Boolean functions from three input to one output. 
In Algorithm \ref{alg2}, as a CPID is given, the global permutation connection is realized. 
It corresponds to a simple one-to-one wiring between hidden and output layers. 
Using the algorithms, a desired PBNN is implemented on the FPGA board. 
Fig. \ref{fg6} (a) shows measured waveform of GBPO with period 50 
corresponding to Fig. \ref{fg4}(a).
Fig. \ref{fg6} (b) shows measured waveform of GBPO with period 100 
corresponding to Fig. \ref{fg4}(b).
The hardware is basic to realize engineering applications such as 
globally stable control signal of switching circuits and 
globally stable approximation signal of time series.

\begin{algorithm}
  \caption{Hidden layer}
  \label{alg1}
  \begin{algorithmic}
    \STATE module HL(parameter $N=17$)
    \STATE (output reg $x[1:N]$,
    \STATE reg $[1:N] x^{t+1}$);
                \STATE wire $rule0,rule1, \cdots, rule7;$
    \STATE parameter CN = 8'b1 ;
    \hfill // Connection Number
    \STATE genvar $j$;
    \STATE for$(j=1; j<=N; j=j+1)$begin
    \STATE \ \ \ \ assign $ rule0=CN[0]^{\ast}($\textasciitilde$x[j-1] \ \&\ $\textasciitilde$x[j] \ \& \ $\textasciitilde$x[j+1]);$
    \hfill // Boolean function
    \STATE \ \ \ \ assign $ rule1=CN[1]^{\ast}($\textasciitilde$x[j-1] \ \&\ $\textasciitilde$x[j] \ \& \ x[j+1]);$
    \STATE \ \ \ \ assign $ rule2=CN[2]^{\ast}($\textasciitilde$x[j-1] \ \&\ x[j] \ \& \ $\textasciitilde$x[j+1]);$
    \STATE \ \ \ \ assign $ rule3=CN[3]^{\ast}($\textasciitilde$x[j-1] \ \&\ x[j] \ \& \ x[j+1]);$
    \STATE \ \ \ \ assign $ rule4=CN[4]^{\ast}(x[j-1] \ \&\ $\textasciitilde$x[j] \ \& \ $\textasciitilde$x[j+1]);$
    \STATE \ \ \ \ assign $ rule5=CN[5]^{\ast}(x[j-1] \ \&\ $\textasciitilde$x[j] \ \& \ x[j+1]);$
    \STATE \ \ \ \ assign $ rule6=CN[6]^{\ast}(x[j-1] \ \&\ x[j] \ \& \ $\textasciitilde$x[j+1]);$
    \STATE \ \ \ \ assign $ rule7=CN[7]^{\ast}(x[j-1] \ \&\ x[j] \ \& \ x[j+1]);$
                \STATE \ \ \ \ assign $x^{t+1}[j] = (rule0)|(rule1)|~...~|(rule7);$
    \STATE end
    \STATE endmodule
  \end{algorithmic}
\end{algorithm}
\begin{algorithm}
  \caption{Output layer}
  \label{alg2}
  \begin{algorithmic}
    \STATE module OL(parameter $N=17$)
    \STATE (input clk$,$load$,$rst$,$
                \STATE input $i[1:N],$
    \STATE output reg $x[1:N]$);
    \STATE reg $[1:N] x^{t+1}$;
    \STATE integer $k$;
    \STATE integer $y[1:N]=$
    \STATE $[1, 3, 11, 14, 4, 13, 8, 15, 12, 7, 16, 10, 5, 17, 6, 2, 9];$
    \STATE // Permutation identifier P1 3 11 14 4 13 8 15 12 7 16 10 5 17 6 2 9
    \STATE always $@$(posedge clk)begin
    \STATE \ \ if(load$==1$)begin
    \STATE \ \ \ \ for$(k=1; k<=N; k=k+1)$begin
    \STATE \ \ \ \ \ \ $x[k] = i[k];$ \hfill // Initial condition
    \STATE \ \ \ \ end
    \STATE \ \ end else if(rst$==1$)begin
    \STATE \ \ \ \ for$(k=1; k<=N; k=k+1)$begin
    \STATE \ \ \ \ \ \ $x[k] = 0;$
    \STATE \ \ \ \ end
    \STATE \ \ end else begin
    \STATE \ \ \ \ for$(k=1; k<=N; k=k+1)$begin
    \STATE \ \ \ \ \ \ $x[k] = x^{t+1}[y[k]];$ \hfill // Permutation
    \STATE \ \ \ \ end
    \STATE \ \ end
    \STATE end
    \STATE HL HL$(x,x^{t+1})$;
    \STATE endmodule
  \end{algorithmic}
\end{algorithm}

\section{Conclusions}
The PBNN and GBPOs are studied in this paper. 
In order to explore GBPOs, we have presented a simple evolutionary algorithm. 
Applying the algorithm to typical examples, 
we have confirmed that the PBNN can have a variety of GBPOs. 
As a first step to realize engineering applications, we have presented FPGA based hardware prototype. 
Using the hardware, typical GBPOs are confirmed experimentally. 
In our future works, 
we should consider various problems including the following. 

1) Computation cost in  the evolutionary algorithm. 

2) Mechanism to reinforce stability of BPOs

3) Engineering applications of GBPOs.

\end{document}